\def\year{2020}\relax
\documentclass[letterpaper]{article} 
\usepackage{aaai20}  
\usepackage{times}  
\usepackage{helvet} 
\usepackage{courier}  
\usepackage[hyphens]{url}  
\usepackage{graphicx} 
\usepackage{graphicx}  
\usepackage{amsmath}

\title{GTC: Guided Training of CTC \\ Towards Efficient and Accurate Scene Text Recognition}
\author{Wenyang Hu,\textsuperscript{\rm 1}\thanks{This work was done when Wenyang was an intern at SenseTime Group Ltd.} Xiaocong Cai,\textsuperscript{\rm 2} Jun Hou,\textsuperscript{\rm 2} Shuai Yi,\textsuperscript{\rm 2} Zhiping Lin\textsuperscript{\rm 1}\\ 
\textsuperscript{\rm 1}Nanyang Technological University\\\textsuperscript{\rm 2}SenseTime Group Ltd. \\ 
huwe0013@e.ntu.edu.sg, \{caixiaocong, houjun, yishuai\}@sensetime.com, ezplin@ntu.edu.sg
}
 
\begin{document}

\maketitle

\begin{abstract}
Connectionist Temporal Classification (CTC) and attention mechanism are two main approaches used in recent scene text recognition works. Compared with attention-based methods, CTC decoder has a much shorter inference time, yet a lower accuracy. To design an efficient and effective model, we propose the guided training of CTC (GTC), where CTC model learns a better alignment and feature representations from a more powerful attentional guidance. With the benefit of guided training, CTC model achieves robust and accurate prediction for both regular and irregular scene text while maintaining a fast inference speed. Moreover, to further leverage the potential of CTC decoder, a graph convolutional network (GCN) is proposed to learn the local correlations of extracted features. Extensive experiments on standard benchmarks demonstrate that our end-to-end model achieves a new state-of-the-art for regular and irregular scene text recognition and needs 6 times shorter inference time than attention-based methods.
\end{abstract}

\section{Introduction}
Scene text recognition has been studied in academia and industry for many years, as it plays an important role in various real-world applications such as vehicle license plate recognition, identity authentication and content analysis. In recent years, many methods proposed \cite{bissacco2013photoocr,shi2016end,shi2018aster} to recognize text in the wild. However, due to different sizes, fonts, colors and character placements of scene texts, scene text recognition is still a challenging task.

\noindent\textbf{Current Recognition Framework} Generally, the framework of scene text recognition models is an encoder-decoder structure. Recent methods mainly use two techniques to train the sequence recognition model, namely Connectionist Temporal Classification (CTC) and attention mechanism. Inspired by speech recognition, CTC is introduced to align the frame-wise probability with labels. In CTC-based methods \cite{liu2016star,shi2016end}, CNN is used to extract feature sequence and the Recurrent Neural Network (RNN) is used to model the feature sequence. They are trained with CTC loss and can make fast prediction using parallel decoding. Attention-based methods use the attention-mechanism to capture the dependencies of each character in a text line, which can learn a better alignment and deeper feature representations than CTC-based methods. Also some rectification methods have been proposed for text image preprocessing. The rectify module transforms the input text image and rectifies character alignments based on Thin-Plate Spline transformation. The rectified images are then passed to the encoder-decoder structure for recognition. This module can be added to CTC or attention-based methods and it is trained in an end-to-end fashion to learn adaptive transformations. 

\begin{figure}[t]
    \centering
    \includegraphics[width=0.45\textwidth]{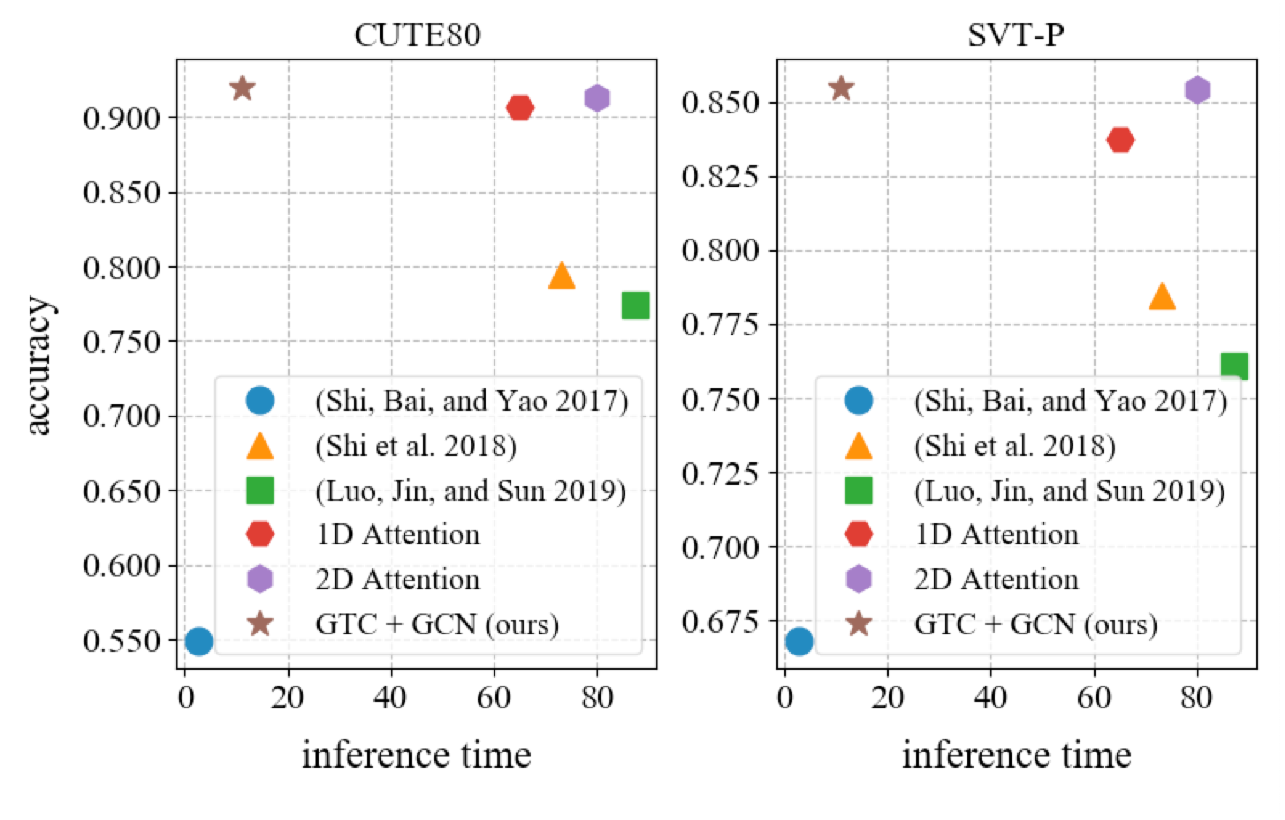}
    \caption{Accuracy versus inference time (ms/image) trade-off tables of different approaches. 1D Attention and 2D Attention represent two attention-based methods. Our method is much faster and more effective.}
    \label{Figure 1}
\end{figure}

\begin{figure*}[ht]
    \centering
    \includegraphics[width=\textwidth]{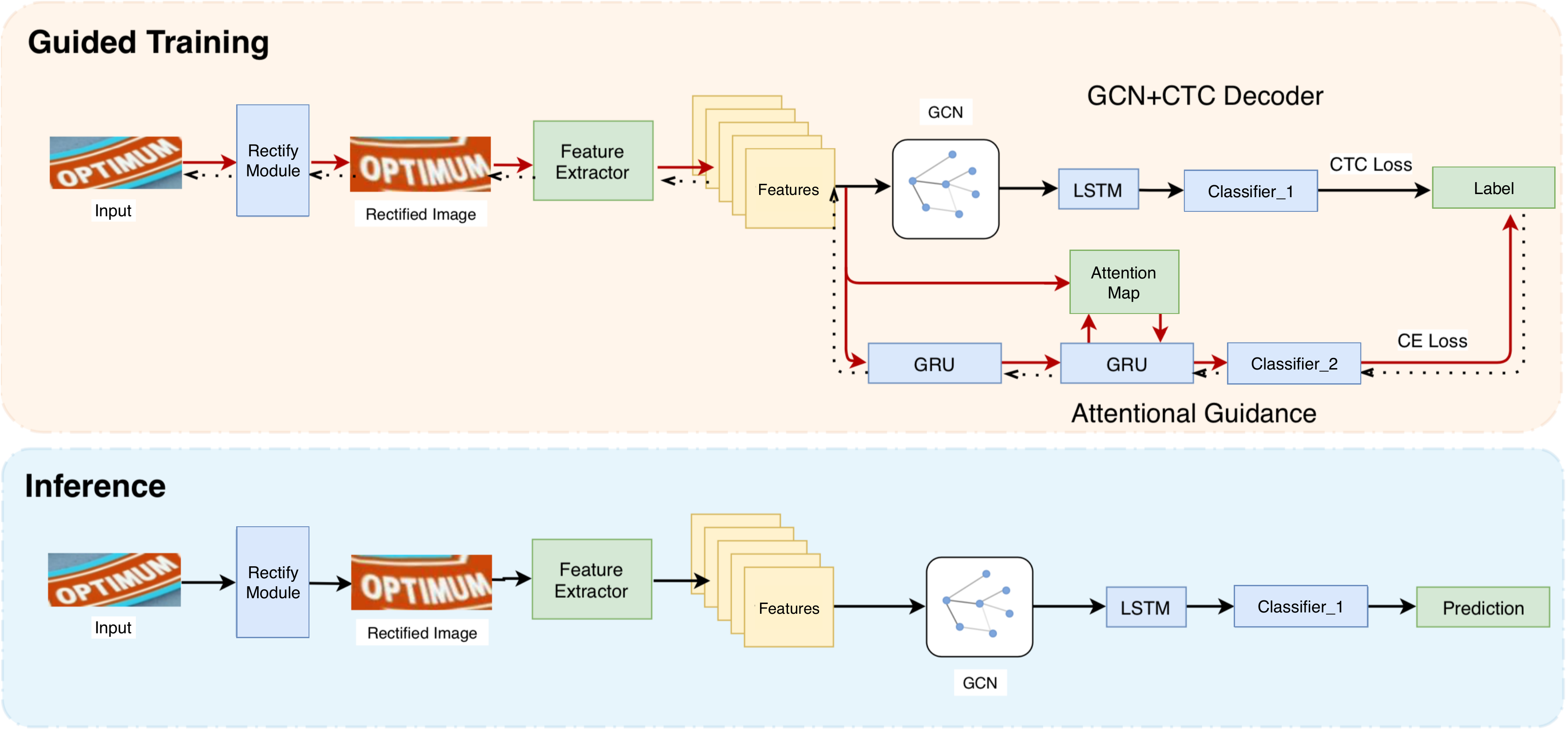}
    \caption{Overview of the proposed method. Different colors of the arrow indicate different computational graphs. Note that the gradients calculated from CTC loss are only used to update the weights of GCN+CTC decoder in the back-propagation process. The ground truth label is the same for both the guidance and GCN+CTC decoder.}
    \label{Figure 2}
\end{figure*}

 \noindent\textbf{Motivations} Though previous approaches give promising results on either regular or irregular scene text recognition, they still have limitations of balancing the trade-off between recognition accuracy and inference time (see Figure 1). As attention-based methods make predictions depending on the features in previous time steps, this non-parallel decoding scheme will slow down the inference process a lot. Although CTC-based methods are relatively efficient, they are not as effective as attention-based methods, where CTC loss misleads the training of its feature alignments and feature representations. With the intention of designing an efficient and effective scene text recognizer, we aim to optimize the CTC model. To overcome the limitations of CTC, we have two motivations: (1) learning better feature representations from a more effective guidance, and (2) building correlations among the local features.

\noindent\textbf{Our Work} We propose the guided training of CTC (GTC), a novel method for general CTC optimization. The proposed method uses an end-to-end trainable framework as shown in Figure 2. The rectify module is a simple transformation network that applies rectification to text images. The CTC decoder is used for both training and evaluation, which leads to efficient inference. For guidance, it is made up of an effective attention decoder which will not be used in the inference. The encoder from a powerful network can learn a better alignment and feature representations, where the feature maps are easier to be decoded. Thus, the CTC model learns from the guidance and becomes more effective. The gradients calculated from cross entropy loss will be used to directly optimize the rectify module, ResNet CNN and the attention decoder, which gives a powerful guidance. The choice of guidance is flexible, which makes this method very general.

As CTC decoding allows repetitions of characters and blank labels, one label can be predicted in multiple time steps. We assume neighbouring time steps have supplementary features and there are certain correlations inside the feature sequence. For example, the letter `H' can be misclassified as `I' if only half part of `H' is considered. To merge the supplementary features of the feature sequence, we further propose a novel GCN module to capture the dependency of each sequence slice and merge the features belonging to the same label based on the correlation. This further improves the robustness and accuracy of the CTC decoder while maintaining a fast inference speed.

In summary, the contributions of this paper are three-fold:

1) We design a novel GTC method for scene text recognition, which is flexible and effective. This method can be adopted to various CTC-based methods and makes the CTC model more effective by learning from powerful guidance, which may become a general method to improve the effectiveness of CTC-based methods. We also use different guidance to show that GTC is a general method for CTC model optimization.

2) This is the first attempt to apply graphs in scene text recognition and build sequence correlations by using GCN, which improves the accuracy and robustness of CTC decoder. Through extensive experiments, we show that the GCN module is effective.

3) Our proposed method has a 6 times shorter inference time than attention-based methods and achieves a new state-of-the-art on most regular and irregular scene text datasets, which is more efficient and effective compared to other works.

\section{Related Work}
\noindent\textbf{Regular Scene Text Recognition} Early works, e.g. \cite{bissacco2013photoocr}, treated scene text recognition as a character segmentation and recognition problem. However, the complicated background and different alignments of scene texts make the character segmentation the most challenging part to be trained. Besides, \cite{jaderberg2015spatial} also used a multi-classification approach to directly predict each word, but this method is heavily constrained by the dictionary size. \cite{lee2016recursive} proposed a recurrent model $R^{2}AM$ with an attention decoder for regular scene text recognition. Inspired by the sequence-to-sequence alignment in speech recognition, \cite{shi2016end} introduced CTC decoder into scene text recognition with a Bidirectional Long Short-Term Memory (BiLSTM) to model the feature sequence, which is known as CRNN. A Gated Recurrent Convolutional Neural Network (GRCNN) was presented by \cite{wang2017gated} which is trained with CTC loss for regular text recognition. Alternatively, \cite{liu2018squeezedtext} proposed a binary convolutional encoder-decoder network (B-CEDNet) trained with cross-entropy loss, and achieved fast inference. However, it is designed for regular scene text recognition and requires pixel-wise labels for training. Inspired by CTC, \cite{bai2018edit} proposed a ``Edit Probability" to optimize the training process, as missing or superfluity of characters may mislead CTC training. \cite{zhang2019sequence} also introduced a domain adaption method to varying length text recognition. The major approach for recent regular text recognition methods is still CTC-based, which enforces the alignment between feature sequence and labels. However, CTC fails on predicting irregular scene text, as the curvatures mislead the alignments.

\noindent\textbf{Irregular Scene Text Recognition} Recognizing irregular scene text has attracted increasing attention in recent years, as it is a more challenging problem. Due to the distortions and curvatures of irregular texts, \cite{shi2018aster} and \cite{shi2016robust} firstly rectified the irregular texts, based on Spatial Transformer Network (STN) \cite{jaderberg2016reading}, which makes text images more regular and easier to be recognized. \cite{luo2019moran} and \cite{zhan2019esir} focused on improving the rectification pipeline to get better transformation results. These methods all use attention-based decoders. CTC-based decoder was also used in \cite{liu2016star}, with a STN rectifying the input image. Instead of rectifying images, some works recognize the irregular text directly. Due to the misalignment between the feature map and the attention region, \cite{cheng2017focusing} used a Focusing Network to adjust the ``attention drift". \cite{liu2018char} also proposed a Character-Aware Neural Network (Char-Net) to rectify the individual characters. However, both methods require character-level annotations. \cite{cheng2018aon} used a multi-direction approach to encode features before the attention network. Alternatively, there are other methods which extend the attention-mechanism into 2D feature maps, as the 2D space captures more spatial dependency. \cite{yang2017learning} and \cite{liao2019scene} recently use 2D local features in their attention networks, but these two works also require character-level annotations. \cite{li2019show} proposed a 2D attention decoder to significantly improve the performance on irregular text recognition, although the inference time is relatively longer. \cite{wang2019simple} proposed a Transformer-based decoder which is also connected to a 2D feature map. This method achieves parallel training, but non-parallel decoding. Among the mentioned works, the attention-based (Attention or STN+Attention) methods generally have higher accuracy on irregular scene text datasets, but the attention decoder slows down the inference. Although STN+CTC maintains a faster inference time, the capacity of CTC decoder constrains its performance.

Different from the mentioned approaches, this paper uses the guided training to optimize the CTC model with better representations of images. We also build spatial and contextual correlations among the sequence by using a designed GCN layer. To our best knowledge, this may be the first work of scene text recognition that uses a guided training method and a GCN layer.

Although \cite{cheng2017focusing} used extensive experiments to show that the direct combination of CTC and attention network does not work well on scene text recognition, they did not give explanations. The reason behind is that CTC degrades the learning of feature representations. We use solid experiments to show that CTC model can achieve a much better performance by learning from an powerful guidance. Different from \cite{kim2017joint} which used a shared encoder by CTC-attention and uses attention decoder for evaluation in speech recognition, encoder and rectification model in our network are solely optimized by the gradients calculated from the guidance. At the test time, only the CTC decoder is used to make predictions and this achieves a much shorter inference time than the attention method. The guidance is only used in the training phase, which supervises the CTC model to learn a better context alignment and feature representations. Therefore, GTC makes the CTC model more efficient and effective.

\section{Methodology}
\subsection{Overview}
Guided training is proposed to overcome the limits of CTC itself. In the inference of CTC model, only the maximum probability of each time step is chosen as the final prediction. However, in training, different probabilities in a single time step contribute to the loss of different CTC paths. The CTC training will make feature representations to tolerate some prediction error. If we have a ground truth label `AB', for a 3 time-steps output, the CTC path (pseudo label) can be `A-B' or `-AB' or `AB-' or `AAB' or `ABB'. As labels for CTC loss calculation are ambiguous, it is confusing to learning feature representations in each time step. Missing or superfluous characters may degrade the learning of its feature alignments and feature representations. Though we used the same encoders for both attention model and CTC model, we found in experiments that the encoder of CTC model has poor feature representations. We assert that the performance of CTC encoder is actually constrained by CTC loss itself. Therefore, a guidance can provide better feature representations for CTC model.

The proposed GTC is described in this section, where a general attention decoder is used as a guidance and our GCN+CTC decoder is used for training and inference. As shown in Figure 2, our network consists of four parts. The first part is a STN which is the same as in \cite{shi2016robust}. It transforms input images into normalized images. The second part is a ResNet backbone for feature extraction, which is widely used in scene text recognition \cite{li2019show,wang2019simple,zhan2019esir}. The third part is an attentional guidance which uses attention mechanism to output the text sequence. The fourth part is a GCN powered CTC decoder which strengthens the correlations of feature sequence. The STN, ResNet-CNN and the attentional guidance are solely trained with cross entropy loss, while the GCN+CTC decoder is trained with CTC loss.

\subsection{Spatial Transformer Network}
As many text images in natural scenes appear with curved texts and different perspectives, the transformation module is adopted for robust and accurate recognition, which applies spatial transformation to text images and normalizes the character region. It is a differentiable module and consists of a localization network and a grid generator. The localization network will predict transformation parameters and use them to create a grid. The grid and the input image will be sampled by the generator to generate the transformed output.

\subsection{Feature Extractor}
We choose ResNet50 \cite{he2016deep} as our network's backbone, which is shown in Table 1. To extract more precise features, we change the original residual block stride from 2 to 1. We also add two max-pooling layers for down-sampling the feature map. The extracted feature sequence $h^{1:T}$ has a fixed height and a varying length, which will be used for decoding.

\begin{table}[t]
\centering
\caption{The configuration of our ResNet50 feature extractor. "Conv" means convolutional layers, provided with its kernel size, output channels, stride and padding. The stride for all `Residual Block' is set to 1. The configuration for "Max-pooling" and "Average-pooling" represents its kernel size, stride, and padding. The overall down-sampling ratio is \textit{W}: 1/4, \textit{H}: 1/16.}
\label{table1}
\begin{tabular}{c|c}
\hline
Layer Name            & Configuration          \\ 
\hline
Conv                  & $7\times7$, $64$, $2\times2$, $3$      \\ 
\hline
Max-pooling           & $3\times3$, $2\times2$, $1$ \\ 
\hline
Residual Block        & 
$\left[
\begin{array}{ccc}
Conv: 1\times 1,  64 \\
Conv: 3\times 3,  64 \\
Conv: 1\times 1,  256 \\
\end{array}
\right] \times 3$\\ 
\hline
Residual Block        & 
$\left[
\begin{array}{ccc}
Conv: 1\times 1,  128 \\
Conv: 3\times 3,  128 \\
Conv: 1\times 1,  512 \\
\end{array}
\right] \times 4$\\ 
\hline
Max-pooling           & $2\times1$, $2\times1$ \\ 
\hline
Residual Block        & 
$\left[
\begin{array}{ccc}
Conv: 1\times 1,  256 \\
Conv: 3\times 3,  256 \\
Conv: 1\times 1,  1024 \\
\end{array}
\right] \times 3$\\ 
\hline
Max-pooling           & $2\times1$, $2\times1$ \\ 
\hline
Residual Block        & 
$\left[
\begin{array}{ccc}
Conv: 1\times 1,  512 \\
Conv: 3\times 3,  512 \\
Conv: 1\times 1,  2048 \\
\end{array}
\right] \times 3$\\ 
\hline
Average-pooling           & $4\times1$, $1\times1$ \\
\hline
\multicolumn{1}{l|}{} & \multicolumn{1}{l}{}  
\end{tabular}
\end{table}

\subsection{Attentional Guidance}
Inspired by machine translation, the sequence-to-sequence model is used to translate a feature sequence into a character sequence, which aligns outputs and labels. The attention mechanism in such a model has the ability to capture output dependencies and focus on character region at each time step. For fair comparisons, we choose a general attention decoder as in \cite{shi2016robust,cheng2017focusing,zhan2019esir} to demonstrate the effectiveness of GTC. We adopt the attentional sequence-to-sequence decoder at the top of the ResNet backbone. It is based on an RNN producing a target sequence of length \textit{T}, denoted by ($y_{1}$, $y_{2}$..., $y_{T}$ ).

The attention decoder either predicts a character or a end-of-sequence
token `EOS'. It stops predicting when it predicts an `EOS'. The Gated Recurrent Cell (GRU) is adopted to learn the attention dependency. At time-step t, output $y_{t}$ is,
\begin{equation}
y_t = Softmax(W^Ts_t),
\end{equation}
where $s_{t}$ is a hidden state of the GRU cell and $W$ is a trainable parameter.

The hidden state $s_{t}$ is updated via the recurrent process of GRU:
\begin{equation}
s_t = GRU(y_{prev},g_t,s_{t-1}),
\end{equation}
where $y_{prev}$ is the embedding vector of the previous output $y_{t-1}$. During training, $y_{t-1}$ is replaced by the ground truth sequence. $g_{t}$ represents the glimpse vector calculated as:
\begin{equation}
g_t = \sum_{i=1}^{T}(\alpha_{ti}h_i),
\end{equation}
where $h_{i}$ is the feature sequence vector of $h^{1:T}$ at the time-step $i$. $\alpha_{t}$ is the attention weight vector as follows:
\begin{equation}
\alpha_{t} = Attention(s_{t-1},h_i),
\end{equation}
which is described in \cite{luong2015effective}.

The attentional guidance is trained with the cross entropy loss and the prediction results will not be used in the evaluation.

\subsection{GCN+CTC Decoder}
Given a sequence of probability distributions $y^{1:T}$ of length $T$, it seeks multiple paths $M^{-1}(l)$ that produce the same label sequence $l$, allowing repetitions of consecutive characters or blank labels $\emptyset$. $y^{t}$ represents the probability distribution at time-step $t$ over the classification labels $L$, where $\emptyset$$\in$$L$. $M$ defines the operation of mapping all possible paths $\pi$ to the target label. For example, it maps the path `-hh-e-ll-l--oo--' into `hello'. CTC trains the network to optimize the summation of probabilities over all paths:

\begin{equation}
p(l | h^{1:T}) = \sum_{\pi\in  M^{-1}(l)}p(\pi | h^{1:T}),
\end{equation}
where the probability of one possible path $\pi$ is calculated as:
\begin{equation}
p(\pi | h^{1:T}) = \prod_{t=1}^{T}{y_{\pi_t}^t},\forall \pi\in M^{-1}(l).
\end{equation}

Based on Equations (5) and (6), CTC trains the network to optimize the loss function:
\begin{equation}
Loss_{CTC} = -log\,p(l | h^{1:T}).
\end{equation}

In CRNN, BiLSTM is used to extract sequence feature by reading the text line from both directions. However, it lack the ability of focusing on local regions, as characters appear in separate locations. Graph Convolutional Networks (GCNs) \cite{kipf2016semi} are an efficient variant of CNNs on graph data, where edges of the graph represent implicit connections within the data. Given a relation defined by a graph, graph convolutions pass the messages from a node to its neighbors. We propose a special GCN layer before the BiLSTM, where a similarity adjacency matrix and a distance matrix are combined to describe the spatial contextual correlations.

Given the feature map $h_{1:T}$ from ResNet CNN, the adjacency matrix is learned by computing pairwise similarity between every two sequence slices:
\begin{equation}
A_{S}(i,j) = f(c_i,c_j),
\end{equation}
The similarity projection function is defined as:
\begin{equation}
f(c_i,c_j) = \frac{c_i\cdot c_j}{||c_i||\,||c_j||},
\end{equation}
where $c_i$ is a linear transformation result of $h_i$. The formula basically calculates pairwise cosine similarities. In addition to using similarity relations to focus on similar features, a distance matrix is also used to constrain the similarity to focus on neighboring features. The distance matrix is defined as:
\begin{equation}
A_{D}(i,j) = \frac{exp(-d_{ij} + \beta)}{exp(-d_{ij} + \beta) +1},
\end{equation}
where $d_{ij} = |i-j|$ and $\beta$ is a scale factor.
Therefore, the final output of our GCN+CTC is calculated as:
\begin{equation}
X = (A_{S}*A_{D})HW_g,
\end{equation}
where $W_g$ is an optional weight matrix. The $X$ is then passed to the BiLSTM for sequence modelling.
\begin{equation}
logits = Seq(X)W_c,
\end{equation}
where $W_c$ is a weight matrix for classification and $Seq$ is the BiLSTM with the hidden size of 512. The logits and label $l$ are finally used to calculate CTC loss to train the GCN+CTC decoder.

In summary, GTC uses a more powerful model to guide CTC decoder, where the gradients calculated from CTC loss will not be used to update the rectify module, ResNet CNN, feature maps or the attentional guidance. CTC Decoder updates itself through the training process of the attentional guidance, where it learns to predict from better feature representations and better alignments. GCN builds correlations among features and further improves the performance.

\section{Experiments}
We conduct experiments on both regular and irregular scene text datasets to evaluate the performance of our proposed method.
\subsection{Datasets}

\noindent \textbf{Synthetic Datasets} There are three public synthetic datasets, namely Synth90K \cite{jaderberg2015spatial}, SynthText \cite{gupta2016synthetic} and SynthAdd \cite{li2019show}. Synth90K is randomly generated based on the 90K common English words, which contains 9-million image instances. In SynthText, texts are randomly blended on full images and text samples are cropped. There are in total 8-million images in SynthText. SynthAdd is also a synthetic dataset with only text line annotations to compensate the lack of special characters. There are 1.6-million images in SynthAdd.

\noindent \textbf{Regular datasets} mainly contains text images with horizontal layout of characters and equal spacing between characters. These images can be simply recognized by reading from left to right.
\begin{quote}
\begin{itemize}

\item\textbf{IIIT5K-Words (IIIT5K)} \cite{mishra2012scene} is collected from Google image searches, which contains 2,000 images for training and 3,000 images for evaluation.

\item\textbf{Street View Text (SVT)} \cite{wang2011end} is collected from Google Street Image, with 257 training images and 647 testing images. Some of these outdoor street images are of low-resolution.

\item\textbf{ICDAR 2003 (IC03)} \cite{lucas2003icdar} is a regular text dataset cropped from real scene text images. It contains 1,156 training images and 1,110 testing images. Instead of filtering the samples which contain non-alphanumeric characters or have fewer than three characters, we use the whole dataset for testing.

\item \textbf{ICDAR 2013 (IC13)} \cite{karatzas2013icdar} has 848 cropped text instances for training and 1095 for testing. It inherits most of IC03's images.

\end{itemize}
\end{quote}

\noindent \textbf{Irregular datasets} contain many hard cases of scene text images. Many of these are curved, rotated and distorted text images.
\begin{quote}
\begin{itemize}

\item\textbf{ICDAR 2015 (IC15)} \cite{karatzas2015icdar} contains 4,468 images for training and 2,077 images for evaluation. These images are cropped from natural images captured by Google Glasses. Thus, many images are blurry, curved and rotated.

\item\textbf{SVT Perspective (SVT-P)} \cite{quy2013recognizing} consists of 645 cropped images, which are from side view images and contain perspective distortions.

\item\textbf{CUTE80} \cite{risnumawan2014robust} contains 288 text patches cropped from natural scene images. Many of these are curved text images of high resolution.

\item\textbf{COCO-Text (COCO)} \cite{veit2016coco} contains 42618 real text images for training and 9837 images for testing.

\end{itemize}
\end{quote}

\begin{table*}[!htb]\centering
\caption{Text line recognition accuracy (in percentages) on public benchmarks, including both regular and irregular datasets. All experiments are compared in a lexicon-free basis. In each column, the state-of-the-art result is shown in \textbf{bold}, and the second best result is shown in \underline{underline}. The methods marked with ``*" are carefully evaluated with rotation strategy. The methods marked with ``+" are trained with both word-level and character-level annotations. Our best model outperforms all the compared methods in the overall recognition rate, and achieves the state-of-the-art on most scene text datasets.}
\vspace{2mm}
\label{table2}
\begin{tabular}{c|c|c|c|c|c|c|c|c|c}
\hline
\multicolumn{2}{c|}{Method} & \multicolumn{4}{|c|}{Regular Text} & \multicolumn{3}{c|}{Irregular Text} & \multicolumn{1}{c}{Infer-Time}   \\ \cline{3-10}
\multicolumn{2}{c|}{}               & IIIT5K   & IC03   & IC13   & SVT  & IC15 & SVT-P & CUTE80 & ms/image \\ \hline

CTC               &  \cite{liu2016star}   & \multicolumn{1}{c|}{83.3}     &    89.9    &    89.1    &   83.6   & -     & 73.5      & -   & -     \\ 
\cline{2-10}  &  \cite{wang2017gated}  &    \multicolumn{1}{c|}{80.8}     &    91.2    &    -    &   81.5   & -     & -      & -   & -     \\
\cline{2-10}     &  \cite{shi2016end}   &    \multicolumn{1}{c|}{81.2}     &    89.9    &    89.6    &   82.7   & -     & 66.8      & 54.9    & 2.7  \\ \hline
\multicolumn{1}{c|}{Attention}  & \cite{lee2016recursive} &    78.4     &    88.7    &    90.0    &   80.7   & -     & -      & -   & -     \\ \cline{2-10}
\multicolumn{1}{c|}{}             & \cite{shi2016robust}   &    81.9     &    90.1    &    88.6    &   81.9   & -     & 71.8      & 59.2   & -     \\ \cline{2-10}
\multicolumn{1}{c|}{}             & \cite{yang2017learning}+   &    -     &    -    &    -    &   -   & -     & 75.8      & 69.3  & -      \\ \cline{2-10}
\multicolumn{1}{c|}{}             & \cite{cheng2017focusing}+   &    87.4     &    94.2    &    93.3    &   85.9   & 70.6     & -      & -   & -     \\ \cline{2-10}
\multicolumn{1}{c|}{}              & \cite{liu2018squeezedtext}+    &    87.0      &    93.1    &    92.9    &   -   & -     & -      & - & -       \\ \cline{2-10}
\multicolumn{1}{c|}{}             & \cite{liu2018char}+   &    92.0     &    92.0    &    91.1    &   85.5   & 74.2     & 78.9      & -   & -     \\ \cline{2-10}
\multicolumn{1}{c|}{}             & \cite{bai2018edit}+   &    88.3     &    94.6    &    \textbf{94.4}    &   87.5   & 73.9     & -      & -   & -     \\ \cline{2-10}
\multicolumn{1}{c|}{}             &  \cite{zhan2019esir}   &    93.3     &    -    &    91.3    &   90.2   & 76.9     & 79.6     & 83.3 & -       \\ \cline{2-10}
\multicolumn{1}{c|}{}              & \cite{shi2018aster}   &    93.4     &    94.5    &    91.8    &   \textbf{93.6}   & 76.1     & 78.5      & 79.5   & 73.1     \\ \cline{2-10}
\multicolumn{1}{c|}{}               & \cite{luo2019moran}   &    91.2     &    95.0    &    92.4    &   88.3   & 68.8     & 76.1      & 77.4    & 87.3    \\ \cline{2-10}
\multicolumn{1}{c|}{}              & \cite{liao2019scene}+    &    91.9     &    -    &    91.5    &   86.4   & -     & -      & 79.9   & -     \\ \cline{2-10}
\multicolumn{1}{c|}{}              &  \cite{li2019show}*    &    95.0     &    -    &    \underline{94.0}    &   91.2   & 78.8     & \textbf{86.4}      & 89.6    & -    \\ \cline{2-10}
\multicolumn{1}{c|}{}              &  \cite{wang2019simple}    &    93.3     &    -    &    91.3    &   88.1   & 77.9     & 80.2      & 85.1    & -    \\ \hline\hline
\multicolumn{1}{c|}{Ours}      & CTC Baseline              &     95.4     &    93.6    &    91.8    &     89.2 & 76.4     & 80.1      & 85.7   & 9.5    \\ \cline{2-10}
\multicolumn{1}{c|}{}      & GTC (1D)              &     \textbf{96.0}     &    \textbf{95.8}    &    93.2    &   91.8   & 79.5     & 85.6      & 91.3     & 10.6  \\ \cline{2-10}
\multicolumn{1}{c|}{}      & GTC (2D)            &     95.0     &    94.6    &    92.6    &   91.2   & 79.3     & 83.4     & 90.6    & 10.6   \\ \cline{2-10}
\multicolumn{1}{c|}{}      & CTC + GCN              &     95.2     &    93.9    &    92.4    &     90.6 & 76.6     & 81.7      & 88.2    & 11.0   \\ \cline{2-10}
\multicolumn{1}{c|}{}      & GTC (1D)  +  GCN       &     95.5     &    95.2    &   94.3     &   \underline{93.2}   & \underline{80.4}     & 85.5      & 92.0   & 11.0    \\ \cline{2-10}
\multicolumn{1}{c|}{}       & GTC (2D)  +  GCN              &     \underline{95.8}     &    \underline{95.5}    &   \textbf{94.4}     &   92.9   & 79.5     & 85.7      & \underline{92.2}   & 11.0    \\ \cline{2-10}
\multicolumn{1}{c|}{}      & GTC (2D)  +  GCN *            &     95.5     &    95.2    &   94.3     &   92.9   & \textbf{82.5}     & \underline{86.2}      & \textbf{92.3}   & -    \\ \hline
\end{tabular}
\end{table*}

\subsection{Implementation Details} We implement our proposed network structure with PyTorch and conduct all experiments on NVIDIA Tesla V100 GPUs with 16GB memory. We use a batch size of 32 on each GPU, with 32 GPUs in total. ADAM optimizer is chosen for training, with the initial learning rate set to $10^{-3}$ and a decay rate of 0.1 every 30000 iterations. We directly train our network using synthetic data (Synth90k, SynthText and SynthAdd) and the training images provided from public benchmarks (IIIT5K, SVT, IC03, IC13, IC15, COCO), which is the same as what is described in (Li et al. 2019). We randomly sample 5.6-million images from those training images for training. The input images are resized to have a fixed height of 64 pixels and a varying length, but not longer than 160 pixels.

During the evaluation, the attentional guidance is abandoned. We directly evaluate the input images by using CTC decoder without any rotation or prediction strategies. Greedy decoding is adopted.

\begin{figure}[!hb]
    \centering
    \includegraphics[width=0.46\textwidth]{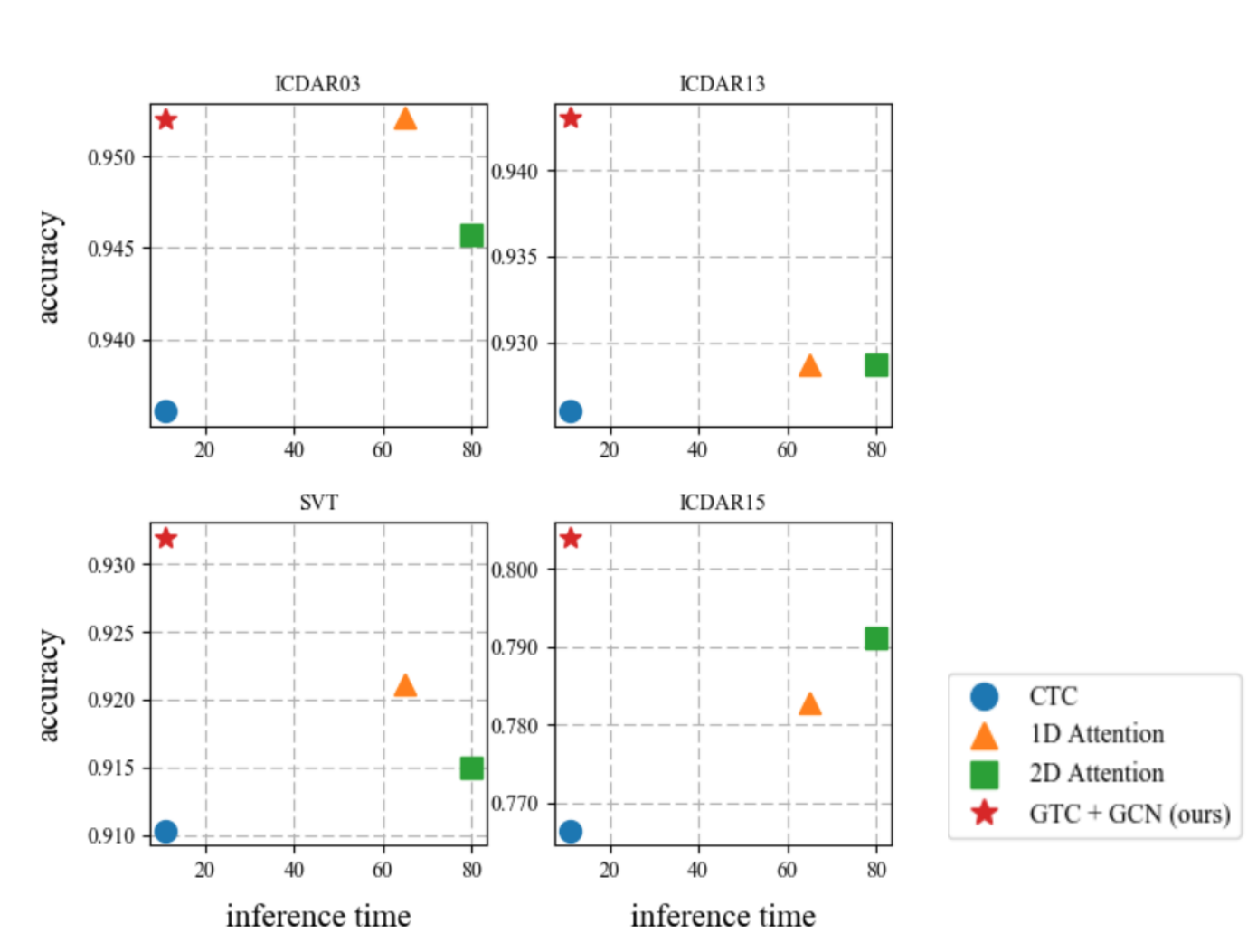}
    \caption{Accuracy versus inference time (ms/image) trade-off tables of different approaches. GTC outperforms CTC and attention-based methods in all datasets.}
    \label{Figure 3}
\end{figure}

\subsection{Experimental Results}
All experiments were evaluated in a lexicon-free condition. We directly evaluated the test images of 7 public datasets and the results are shown in Table 2. `1D' denotes the attentional guidance described in the Methodology. We also evaluated GTC by using another attentional guidance from \cite{li2019show}, which shows the robustness of GTC. The related experimental results are denoted by `2D'. \cite{li2019show} rotated testing images by 90 degrees clockwise and anticlockwise respectively, and recognized them together with the original image. For fair comparison, our experiment in the last row of Table 2 also used the rotation strategy. Our other experiments directly evaluated the test images without rotation. Therefore, we claim that we achieved the state-of-the-art by using only word annotations and public datasets. 

\begin{table}[t]
\centering
\caption{Comparison between GTC and guided training of attention. The performance was measured by using overall accuracy on all public benchmarks.}
\label{table4}
\begin{tabular}{|c|c|}
\hline
Experiment            &   overall accuracy     \\ \hline
Use CTC to Guide Attention                 &    87.87\%    \\ \hline
 Use Attention to Guide CTC (GTC)           &  90.06\%      \\ \hline
\end{tabular}
\end{table}

Our GTC method outperforms CTC-based methods a lot while it maintains a fast inference. The GCN module has also been verified to model a better feature sequence for irregular texts. Note that all GTC results are based on the predictions of CTC decoder.

To evaluate the efficiency of our approach, we also conduct experiments to analyze the inference time of different methods. We fix the batch size as 1 and run all test experiments on the same device. The inference time is measured on a single NVIDIA Titan X GPU with 12 GB memory. The comparison of our method and other methods is shown in Figure 3. The results show that our GTC method achieves 6 times faster in the inference than attention-based methods and maintains the highest recognition rate. The GCN+CTC decoder is even 14+ times faster than the attention decoder. The inference time is 54 ms/image for attention decoder and 3.7 ms/image for the GCN+CTC decoder.

\subsection{Ablation Study}

In this experiment, CTC is used to guide the attention decoder instead. The encoder is trained with the CTC loss and the attention decoder is used for evaluation. The result in Table 3 shows that CTC is not an effective guidance compared with Attention. The result also indicates that the CTC loss harms the training process and produces poor feature representations. Therefore, this experiment suggests that a powerful guidance is necessary in the guided training in order to improve the overall performance.

Besides, we find that GTC also has better transformation results compared with STN+CTC framework (see Figure 4). As attention mechanism is more sensitive to spatial information, the transformation module can be more responsive.

To clearly describe how GCN works, we present visualizations in Figure 5. With distance matrix and similarity matrix multiplied point-wisely, we get the adjacency matrix of GCN, which focuses on local similar features. Thus, it establishes the local correlations for better sequence modelling. 
\begin{figure}[ht]
    \centering
    \includegraphics[width=0.45\textwidth]{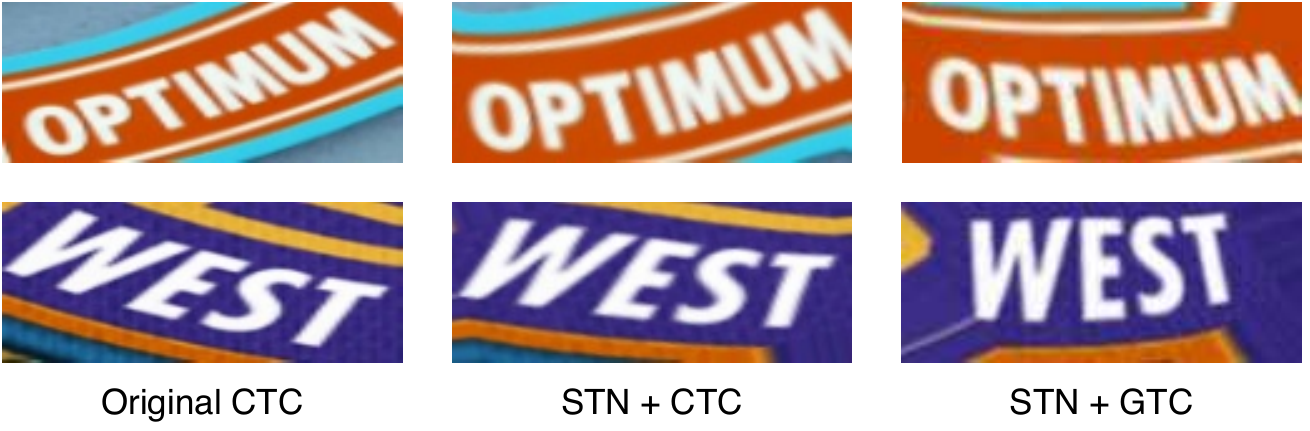}
    \caption{Examples of original images and rectified images from different approaches.}
    \label{Figure 4}
\end{figure}

\begin{figure}[!htb]
    \centering
    \includegraphics[width=0.45\textwidth]{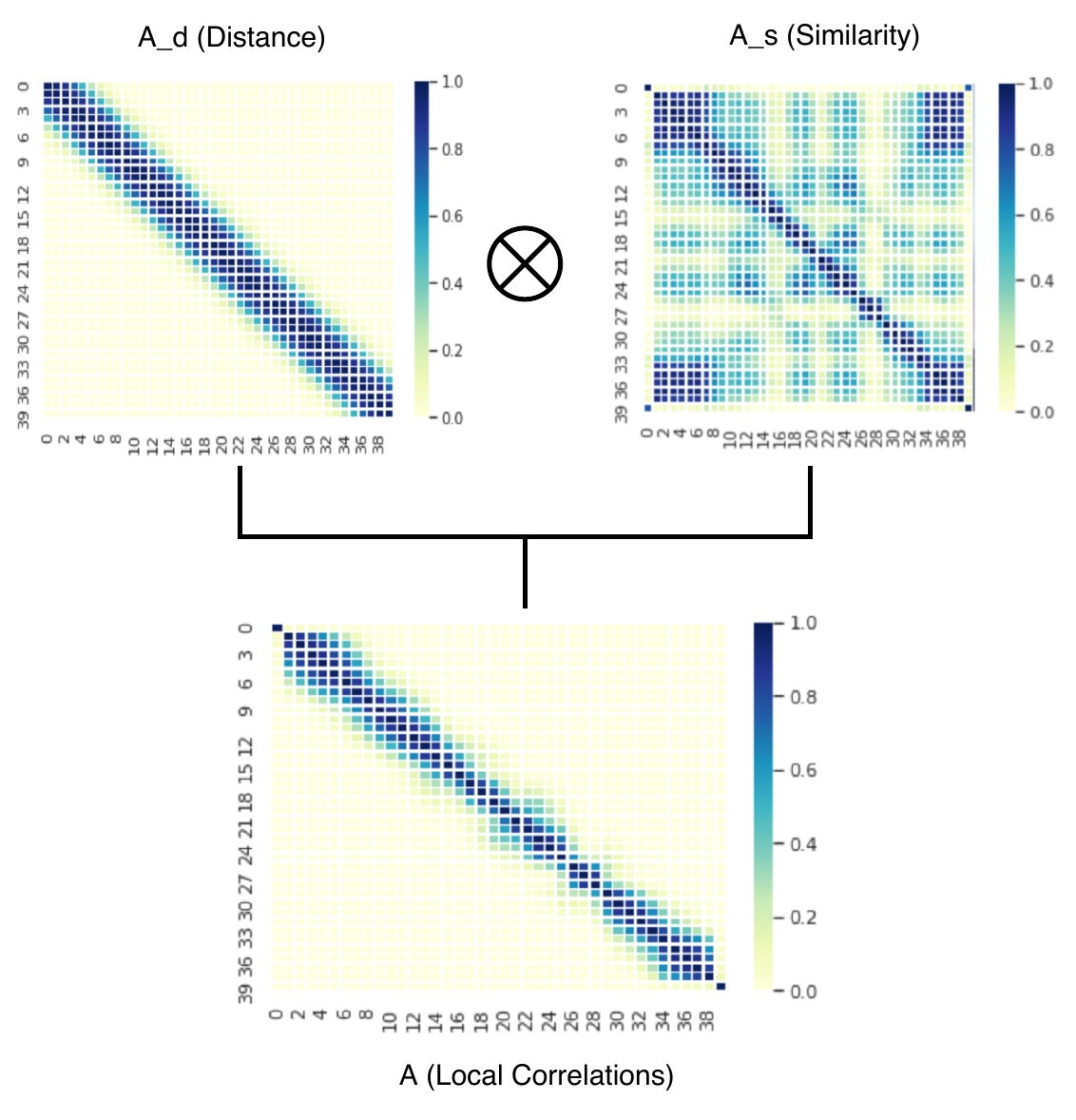}
    \caption{Visualization of the similarity matrix, the distance matrix and their point-wise product.}
    \label{Figure 5}
\end{figure}

\section{Conclusion}
In this paper, we propose an effective and efficient GTC method to significantly improve the robustness and performance on scene text recognition, which overcomes the limitations of CTC. This work is the first attempt to use GCN to learn the local correlations of feature sequences, which further improves CTC performance. We conduct experiments to evaluate the effectiveness of our method on 7 public benchmarks and the method achieves a new state-of-the-art on most datasets.

\bibliography{gtc.bib}

\begin{thebibliography}{}

\bibitem[\protect\citeauthoryear{Bai \bgroup et al\mbox.\egroup
  }{2018}]{bai2018edit}
Bai, F.; Cheng, Z.; Niu, Y.; Pu, S.; and Zhou, S.
\newblock 2018.
\newblock Edit probability for scene text recognition.
\newblock In {\em Proceedings of the IEEE Conf. Comp. Vis. Patt. Recogn},
  1508--1516.

\bibitem[\protect\citeauthoryear{Bissacco \bgroup et al\mbox.\egroup
  }{2013}]{bissacco2013photoocr}
Bissacco, A.; Cummins, M.; Netzer, Y.; and Neven, H.
\newblock 2013.
\newblock Photoocr: Reading text in uncontrolled conditions.
\newblock In {\em Proceedings of the IEEE International Conf. Comp. Vis.},
  785--792.

\bibitem[\protect\citeauthoryear{Cheng \bgroup et al\mbox.\egroup
  }{2017}]{cheng2017focusing}
Cheng, Z.; Bai, F.; Xu, Y.; Zheng, G.; Pu, S.; and Zhou, S.
\newblock 2017.
\newblock Focusing attention: Towards accurate text recognition in natural
  images.
\newblock In {\em Proceedings of the IEEE International Conf. Comp. Vis.},
  5076--5084.

\bibitem[\protect\citeauthoryear{Cheng \bgroup et al\mbox.\egroup
  }{2018}]{cheng2018aon}
Cheng, Z.; Xu, Y.; Bai, F.; Niu, Y.; Pu, S.; and Zhou, S.
\newblock 2018.
\newblock Aon: Towards arbitrarily-oriented text recognition.
\newblock In {\em Proceedings of the IEEE Conf. Comp. Vis. Patt. Recogn},
  5571--5579.

\bibitem[\protect\citeauthoryear{Gupta, Vedaldi, and
  Zisserman}{2016}]{gupta2016synthetic}
Gupta, A.; Vedaldi, A.; and Zisserman, A.
\newblock 2016.
\newblock Synthetic data for text localisation in natural images.
\newblock In {\em Proceedings of the IEEE Conf. Comp. Vis. Patt. Recogn},
  2315--2324.

\bibitem[\protect\citeauthoryear{He \bgroup et al\mbox.\egroup
  }{2016}]{he2016deep}
He, K.; Zhang, X.; Ren, S.; and Sun, J.
\newblock 2016.
\newblock Deep residual learning for image recognition.
\newblock In {\em Proceedings of the IEEE Conf. Comp. Vis. Patt. Recogn},
  770--778.

\bibitem[\protect\citeauthoryear{Jaderberg \bgroup et al\mbox.\egroup
  }{2015}]{jaderberg2015spatial}
Jaderberg, M.; Simonyan, K.; Zisserman, A.; et~al.
\newblock 2015.
\newblock Spatial transformer networks.
\newblock In {\em Advances in neural information processing systems},
  2017--2025.

\bibitem[\protect\citeauthoryear{Jaderberg \bgroup et al\mbox.\egroup
  }{2016}]{jaderberg2016reading}
Jaderberg, M.; Simonyan, K.; Vedaldi, A.; and Zisserman, A.
\newblock 2016.
\newblock Reading text in the wild with convolutional neural networks.
\newblock {\em International Journal of Computer Vision} 116(1):1--20.

\bibitem[\protect\citeauthoryear{Karatzas \bgroup et al\mbox.\egroup
  }{2013}]{karatzas2013icdar}
Karatzas, D.; Shafait, F.; Uchida, S.; Iwamura, M.; i~Bigorda, L.~G.; Mestre,
  S.~R.; Mas, J.; Mota, D.~F.; Almazan, J.~A.; and De~Las~Heras, L.~P.
\newblock 2013.
\newblock Icdar 2013 robust reading competition.
\newblock In {\em 2013 12th International Conference on Document Analysis and
  Recognition},  1484--1493.
\newblock IEEE.

\bibitem[\protect\citeauthoryear{Karatzas \bgroup et al\mbox.\egroup
  }{2015}]{karatzas2015icdar}
Karatzas, D.; Gomez-Bigorda, L.; Nicolaou, A.; Ghosh, S.; Bagdanov, A.;
  Iwamura, M.; Matas, J.; Neumann, L.; Chandrasekhar, V.~R.; Lu, S.; et~al.
\newblock 2015.
\newblock Icdar 2015 competition on robust reading.
\newblock In {\em 2015 13th International Conference on Document Analysis and
  Recognition},  1156--1160.
\newblock IEEE.

\bibitem[\protect\citeauthoryear{Kim, Hori, and Watanabe}{2017}]{kim2017joint}
Kim, S.; Hori, T.; and Watanabe, S.
\newblock 2017.
\newblock Joint ctc-attention based end-to-end speech recognition using
  multi-task learning.
\newblock In {\em 2017 IEEE ICASSP},  4835--4839.
\newblock IEEE.

\bibitem[\protect\citeauthoryear{Kipf and Welling}{2016}]{kipf2016semi}
Kipf, T.~N., and Welling, M.
\newblock 2016.
\newblock Semi-supervised classification with graph convolutional networks.
\newblock {\em arXiv preprint arXiv:1609.02907}.

\bibitem[\protect\citeauthoryear{Lee and Osindero}{2016}]{lee2016recursive}
Lee, C.-Y., and Osindero, S.
\newblock 2016.
\newblock Recursive recurrent nets with attention modeling for ocr in the wild.
\newblock In {\em Proceedings of the IEEE Conf. Comp. Vis. Patt. Recogn},
  2231--2239.

\bibitem[\protect\citeauthoryear{Li \bgroup et al\mbox.\egroup
  }{2019}]{li2019show}
Li, H.; Wang, P.; Shen, C.; and Zhang, G.
\newblock 2019.
\newblock Show, attend and read: A simple and strong baseline for irregular
  text recognition.
\newblock In {\em Proceedings of the AAAI Conference on Artificial
  Intelligence}, volume~33,  8610--8617.

\bibitem[\protect\citeauthoryear{Liao \bgroup et al\mbox.\egroup
  }{2019}]{liao2019scene}
Liao, M.; Zhang, J.; Wan, Z.; Xie, F.; Liang, J.; Lyu, P.; Yao, C.; and Bai, X.
\newblock 2019.
\newblock Scene text recognition from two-dimensional perspective.
\newblock In {\em Proceedings of the AAAI Conference on Artificial
  Intelligence}, volume~33,  8714--8721.

\bibitem[\protect\citeauthoryear{Liu \bgroup et al\mbox.\egroup
  }{2016}]{liu2016star}
Liu, W.; Chen, C.; Wong, K.-Y.~K.; Su, Z.; and Han, J.
\newblock 2016.
\newblock Star-net: a spatial attention residue network for scene text
  recognition.
\newblock In {\em BMVC}, volume~2, ~7.

\bibitem[\protect\citeauthoryear{Liu \bgroup et al\mbox.\egroup
  }{2018}]{liu2018squeezedtext}
Liu, Z.; Li, Y.; Ren, F.; Goh, W.~L.; and Yu, H.
\newblock 2018.
\newblock Squeezedtext: A real-time scene text recognition by binary
  convolutional encoder-decoder network.
\newblock In {\em Thirty-Second AAAI Conference on Artificial Intelligence}.

\bibitem[\protect\citeauthoryear{Liu, Chen, and Wong}{2018}]{liu2018char}
Liu, W.; Chen, C.; and Wong, K.-Y.~K.
\newblock 2018.
\newblock Char-net: A character-aware neural network for distorted scene text
  recognition.
\newblock In {\em Thirty-Second AAAI Conference on Artificial Intelligence}.

\bibitem[\protect\citeauthoryear{Lucas \bgroup et al\mbox.\egroup
  }{2003}]{lucas2003icdar}
Lucas, S.~M.; Panaretos, A.; Sosa, L.; Tang, A.; Wong, S.; and Young, R.
\newblock 2003.
\newblock Icdar 2003 robust reading competitions.
\newblock In {\em Seventh International Conference on Document Analysis and
  Recognition, 2003. Proceedings.},  682--687.
\newblock Citeseer.

\bibitem[\protect\citeauthoryear{Luo, Jin, and Sun}{2019}]{luo2019moran}
Luo, C.; Jin, L.; and Sun, Z.
\newblock 2019.
\newblock Moran: A multi-object rectified attention network for scene text
  recognition.
\newblock {\em Pattern Recognition} 90:109--118.

\bibitem[\protect\citeauthoryear{Luong, Pham, and
  Manning}{2015}]{luong2015effective}
Luong, M.-T.; Pham, H.; and Manning, C.~D.
\newblock 2015.
\newblock Effective approaches to attention-based neural machine translation.
\newblock {\em arXiv preprint arXiv:1508.04025}.

\bibitem[\protect\citeauthoryear{Mishra, Alahari, and
  Jawahar}{2012}]{mishra2012scene}
Mishra, A.; Alahari, K.; and Jawahar, C.
\newblock 2012.
\newblock Scene text recognition using higher order language priors.
\newblock In {\em Proc. British Mach. Vis. Conf},  1--11.

\bibitem[\protect\citeauthoryear{Quy~Phan \bgroup et al\mbox.\egroup
  }{2013}]{quy2013recognizing}
Quy~Phan, T.; Shivakumara, P.; Tian, S.; and Lim~Tan, C.
\newblock 2013.
\newblock Recognizing text with perspective distortion in natural scenes.
\newblock In {\em Proceedings of the IEEE International Conf. Comp. Vis.},
  569--576.

\bibitem[\protect\citeauthoryear{Risnumawan \bgroup et al\mbox.\egroup
  }{2014}]{risnumawan2014robust}
Risnumawan, A.; Shivakumara, P.; Chan, C.~S.; and Tan, C.~L.
\newblock 2014.
\newblock A robust arbitrary text detection system for natural scene images.
\newblock {\em Expert Systems with Applications} 41(18):8027--8048.

\bibitem[\protect\citeauthoryear{Shi, Bai, and Yao}{2016}]{shi2016end}
Shi, B.; Bai, X.; and Yao, C.
\newblock 2016.
\newblock An end-to-end trainable neural network for image-based sequence
  recognition and its application to scene text recognition.
\newblock {\em IEEE transactions on pattern analysis and machine intelligence}
  39(11):2298--2304.

\bibitem[\protect\citeauthoryear{Shi \bgroup et al\mbox.\egroup
  }{2016}]{shi2016robust}
Shi, B.; Wang, X.; Lyu, P.; Yao, C.; and Bai, X.
\newblock 2016.
\newblock Robust scene text recognition with automatic rectification.
\newblock In {\em Proceedings of the IEEE Conf. Comp. Vis. Patt. Recogn},
  4168--4176.

\bibitem[\protect\citeauthoryear{Shi \bgroup et al\mbox.\egroup
  }{2018}]{shi2018aster}
Shi, B.; Yang, M.; Wang, X.; Lyu, P.; Yao, C.; and Bai, X.
\newblock 2018.
\newblock Aster: An attentional scene text recognizer with flexible
  rectification.
\newblock {\em IEEE transactions on pattern analysis and machine intelligence}.

\bibitem[\protect\citeauthoryear{Veit \bgroup et al\mbox.\egroup
  }{2016}]{veit2016coco}
Veit, A.; Matera, T.; Neumann, L.; Matas, J.; and Belongie, S.
\newblock 2016.
\newblock Coco-text: Dataset and benchmark for text detection and recognition
  in natural images.
\newblock {\em arXiv preprint arXiv:1601.07140}.

\bibitem[\protect\citeauthoryear{Wang and Hu}{2017}]{wang2017gated}
Wang, J., and Hu, X.
\newblock 2017.
\newblock Gated recurrent convolution neural network for ocr.
\newblock In {\em Advances in Neural Information Processing Systems},
  335--344.

\bibitem[\protect\citeauthoryear{Wang, Babenko, and
  Belongie}{2011}]{wang2011end}
Wang, K.; Babenko, B.; and Belongie, S.
\newblock 2011.
\newblock End-to-end scene text recognition.
\newblock In {\em 2011 International Conf. Comp. Vis.},  1457--1464.
\newblock IEEE.

\bibitem[\protect\citeauthoryear{Wang \bgroup et al\mbox.\egroup
  }{2019}]{wang2019simple}
Wang, P.; Yang, L.; Li, H.; Deng, Y.; Shen, C.; and Zhang, Y.
\newblock 2019.
\newblock A simple and robust convolutional-attention network for irregular
  text recognition.
\newblock {\em arXiv preprint arXiv:1904.01375}.

\bibitem[\protect\citeauthoryear{Yang \bgroup et al\mbox.\egroup
  }{2017}]{yang2017learning}
Yang, X.; He, D.; Zhou, Z.; Kifer, D.; and Giles, C.~L.
\newblock 2017.
\newblock Learning to read irregular text with attention mechanisms.
\newblock In {\em IJCAI}, volume~1, ~3.

\bibitem[\protect\citeauthoryear{Zhan and Lu}{2019}]{zhan2019esir}
Zhan, F., and Lu, S.
\newblock 2019.
\newblock Esir: End-to-end scene text recognition via iterative image
  rectification.
\newblock In {\em Proceedings of the IEEE Conf. Comp. Vis. Patt. Recogn},
  2059--2068.

\bibitem[\protect\citeauthoryear{Zhang \bgroup et al\mbox.\egroup
  }{2019}]{zhang2019sequence}
Zhang, Y.; Nie, S.; Liu, W.; Xu, X.; Zhang, D.; and Shen, H.~T.
\newblock 2019.
\newblock Sequence-to-sequence domain adaptation network for robust text image
  recognition.
\newblock In {\em Proceedings of the IEEE Conf. Comp. Vis. Patt. Recogn},
  2740--2749.

\end{thebibliography}
\bibliographystyle{aaai}

\end{document}